\documentclass[10pt,twocolumn,letterpaper]{article}

\usepackage{cvpr}              

\usepackage{graphicx}
\usepackage{amsmath}
\usepackage{amssymb}
\usepackage{booktabs}

\usepackage[pagebackref,breaklinks,colorlinks]{hyperref}

\usepackage[capitalize]{cleveref}
\crefname{section}{Sec.}{Secs.}
\Crefname{section}{Section}{Sections}
\Crefname{table}{Table}{Tables}
\crefname{table}{Tab.}{Tabs.}

\def\cvprPaperID{*****} 
\def\confName{CVPR}
\def\confYear{2023}

\begin{document}

\title{Exploring Large-scale Unlabeled Faces to Enhance Facial Expression Recognition}

\author{
Jun Yu, Zhongpeng Cai, Renda Li, Gongpeng Zhao, Guochen Xie, Jichao Zhu, Wangyuan Zhu\\
University of Science and Technology of China\\
\tt\small harryjun@ustc.edu.cn\\
\tt\small \{zpcai,rdli,zgp0531,xiegc,jichaozhu,zhuwangyuan\}@mail.ustc.edu.cn
}

\maketitle

\begin{abstract}
Facial Expression Recognition (FER) is an important task in computer vision and has wide applications in human-computer interaction, intelligent security, emotion analysis, and other fields. However, the limited size of FER datasets limits the generalization ability of expression recognition models, resulting in ineffective model performance. To address this problem, we propose a semi-supervised learning framework that utilizes unlabeled face data to train expression recognition models effectively. Our method uses a dynamic threshold module (\textbf{DTM}) that can adaptively adjust the confidence threshold to fully utilize the face recognition (FR) data to generate pseudo-labels, thus improving the model's ability to model facial expressions. In the ABAW5 EXPR task, our method achieved excellent results on the official validation set.
\end{abstract}

\section{Introduction}
\label{sec:intro}

According to psychology Research\cite{mehrabian1974approach} by scientist A.Mehrabia shows that in human daily communication, the information transmitted through language only accounts for 7\% of the total information, while the information transmitted through facial expressions reaches 55\% of the total information. Therefore, it is significant to build a robust Facial Expression Recognition (FER) System. In recent years, many FER methods\cite{wen2021distract, she2021dive, li2022crs, li2022towards, zeng2022face2exp, lukov2022teaching, zhang2022learn,chang2021learning} achieved state-of-the-art performance on several benchmark datasets (\textit{e.g.} RAF-DB\cite{li2017reliable},  SFEW\cite{yu2015image} and AffectNet\cite{mollahosseini2017affectnet}). 

In order to promote the development of facial expression recognition, Kollias et al. \cite{zafeiriou2017aff,kollias2019deep, kollias2019face, kollias2019expression, kollias2020analysing,kollias2021affect,kollias2021analysing,kollias2021distribution,kollias2022abaw, kollias2023abaw,kollias2023abaw2} organize the competition on Affective Behavior Analysis in-the-wild (ABAW). The 5th Workshop and Competition on Affective Behavior Analysis in-the-wild (ABAW), will be held in conjunction with the IEEE Computer Vision and Pattern Recognition Conference (CVPR), 2023. 


\begin{figure}[h]
    \centering
    \begin{subfigure}{0.8\linewidth}
        \centering
        \includegraphics[width=\linewidth]{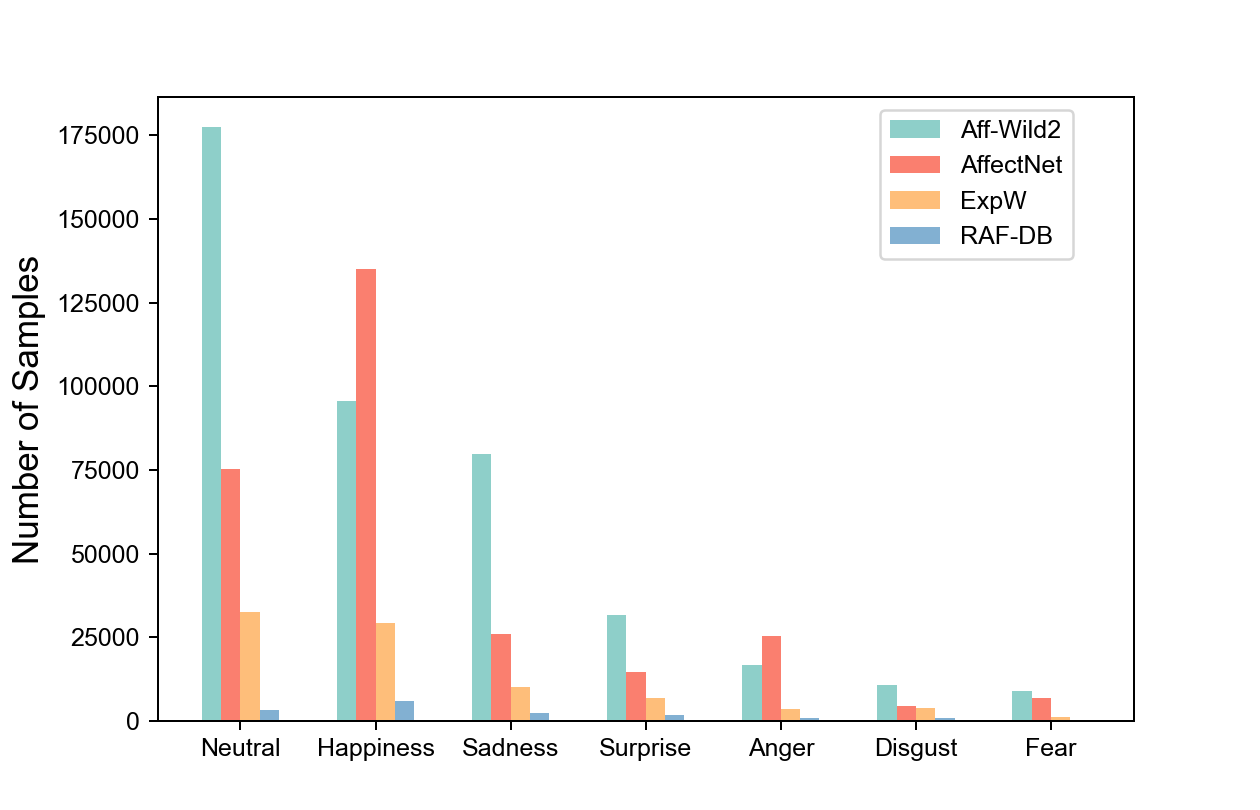}
        \caption{}
        \label{fig:a}
    \end{subfigure}
    \hspace{0.05\linewidth}
    \begin{subfigure}{0.8\linewidth}
        \centering
        \includegraphics[width=\linewidth]{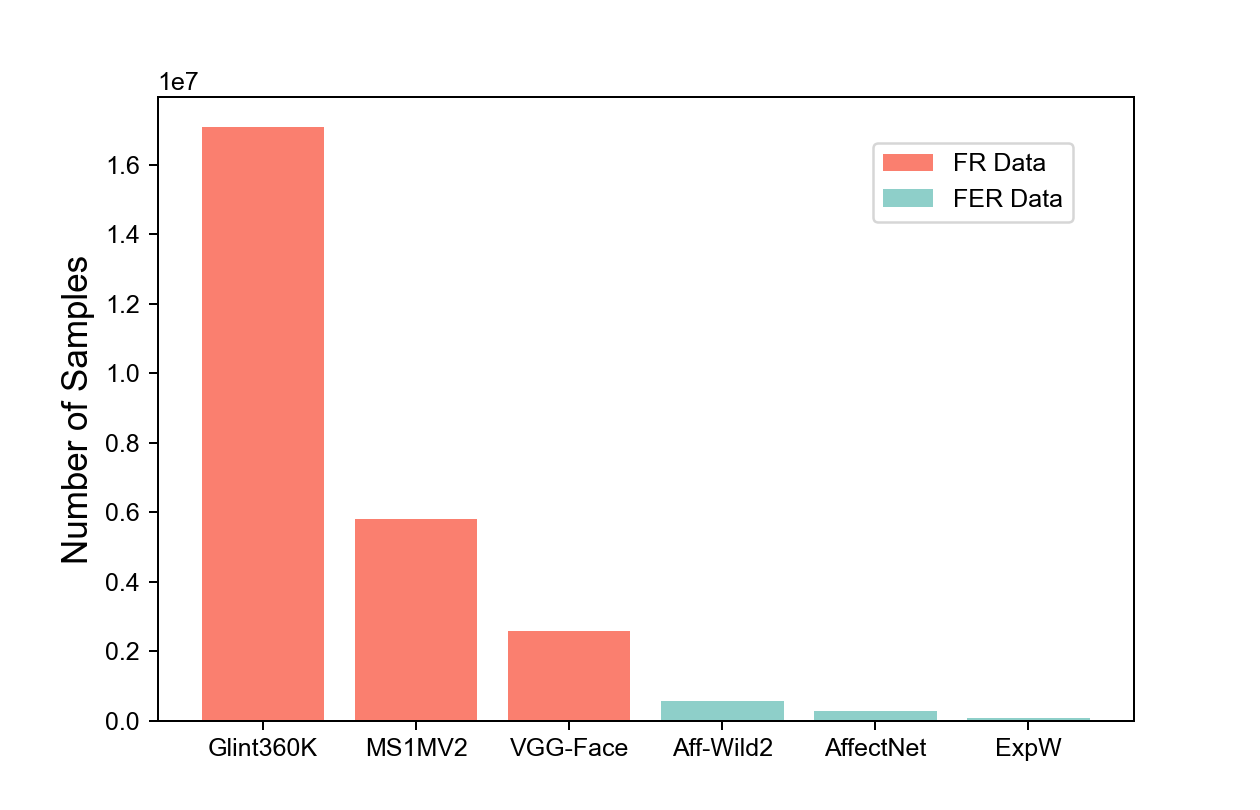}
        \caption{}
        \label{fig:b}
    \end{subfigure}
    \caption{(a) The class distribution of FER datasets. (b) The number of samples of FER data and FR data.}
    \label{fig:Motivation}
\end{figure}

In traditional fully supervised face expression recognition methods \cite{2004Feature,2005Feature,2006Face,2010Expression,tian2011facial,chen2021understanding,chang2021learning,xue2021transfer}, The accuracy of model predictions relies heavily on a large amount of high-quality labeled data. As shown in Figure \ref{fig:Motivation} (a), existing FER training datasets are biased towards some majority classes, which leads to poor test accuracy for the minority classes. The lowest number of classes is less than one-tenth of the highest number of classes. This seriously affects the overall performance of the model. As we all know, it is expensive to obtain a large scale labeled FER data, which makes it difficult to expand the FER training datasets.
However, as shown in \ref{fig:Motivation} (b), the size of Face Recognition (FR) data is much larger than that of FER data. Thus, how to remove the inconsistent data distribution between FR data and FER data becomes an urgent problem of utilizing the (FR) data to enhance  Facial Expression Recognition models.

In this paper, we adopted a semi-supervised approach to obtain pseudo-labels for unlabeled data, in order to obtain sufficient training data to help the model to extract facial expressions. At the same time, to alleviate the problem of class-imbalanced dataset, we uniformly sampled the labeled facial expression samples to correct the bias learned by the model from the unlabeled data. We consider that a fixed threshold cannot fully utilize the data and cannot adapt to the class-imbalanced data. Moreover, considering that the discriminative ability of the model can be significantly improved with the increase of training steps, we designed a dynamic threshold module (\textbf{DTM}) that can adjust the confidence threshold with different classes and training steps to fully utilize the data.

To sum up, our contributions  can be summarized as:
\begin{itemize}
    \item We have built a semi-supervised learning framework for the task of facial expression recognition. It can apply the FR (face recognition) data to the task of facial expression recognition through the use of pseudo-labels, which greatly alleviates the problem of small facial expression datasets.
    \item We designed a dynamic threshold module (\textbf{DTM}) for the Semi-Supervised. It can dynamically adjust the confidence threshold for different stages of training and different expression categories, to fully utilize the FR data to generate pseudo-labels.
    \item In the ABAW5 competition, our method achieve great performance on official validation set.
\end{itemize}

\section{Related Work}
\label{sec:Relatedwork}

\subsection{Facial Expression Recognition}

The facial expression recognition task is a classic task in the field of pattern recognition, and the approach\cite{wen2021distract, she2021dive,  zhang2022learn,chen2021understanding,chang2021learning,xue2021transfer} of making full use of fully supervised data once gained very great progress in the field of FER. In recent years, attention has been focused on extending the dataset to obtain larger scale datasets. To solve the label confusion problem between different expression recognition datasets. IPA2LT \cite{zeng2018facial} is the first work to address the annotation inconsistency in different facial expression datasets. They proposed LTNet embedded with a scheme of discovering the latent truth from multiple inconsistent labels and the input images. Ada-CM \cite{li2022towards} is the first solution to explore the dynamic confidence margin in Semi-Supervised Deep Facial Expression Recognition. They designed an adaptive confidence margin to dynamically learn on all unlabeled data for the model’s training and conducted a feature-level contrastive objective to learn effective features by applying the InfoNCE\cite{wu2021rethinking} loss. Face2Exp \cite{zeng2022face2exp} proposed the Meta-Face2Exp framework to extract de-biased knowledge from auxiliary FR data through the meta optimization framework. utilized unlabeled face data to enhance expression recognition through the meta optimization framework.

\subsection{Learning with Unlabeled Data}
An important direction for learning methods using unlabeled data is semi-supervised learning. A popular class of Semi-Supervised Learning methods is to generate an artificial label for an unlabeled image and train the model to predict that artificial label on the input of an unlabeled image \cite{lee2013pseudo,xie2020self}. Similarly, consistent regularization \cite{bachman2014learning,laine2016temporal,sajjadi2016regularization} uses the predictive distribution of the model to obtain an artificial label after randomly modifying the input or model function. An artificial label is obtained using the predictive distribution of the model after randomly modifying the input or the model function.
Fixmatch\cite{sohn2020fixmatch} combines the advantages of these two methods, which use weak data augmentation and strong data augmentation of samples to achieve consistent regularization and obtain pseudo-labeled data by samples with confidence levels greater than a threshold. But the problem of this approach is that his modeling ability is limited in the early stage of model training because the threshold value is fixed.
To solve this problem, Flexmatch\cite{zhang2021flexmatch} proposed Curriculum Pseudo Labeling (CPL), a curriculum learning approach to leverage unlabeled data according to the model’s learning status. It can flexibly adjust thresholds for different classes at each time step to let pass informative unlabeled data and their pseudo labels. In addition, Dash[] performs selection by retaining only samples with losses less than a given threshold in each update iteration, which is dynamically adjusted through iterations.

\subsection{Affective Behavior Analysis in-the-wild}
Zhang \emph{et al.} \cite{zhang2022transformer}utilized the multimodal information from the images, audio and text and proposed a unified multimodal framework to fully use the emotion information, which achieved the best performance in ABAW3 competition. Jeong \emph{et al.} \cite{jeong2022classification}proposed a multi-head cross attention networks and pretrained on Glint360K\cite{an2021partial} and some private commercial datasets. Xue \emph{et al.} \cite{xue2022coarse} proposed the Coarse-to-Fine Cascaded networks (CFC) to address the label ambiguity problem and used smooth predicting method to post-process the extracted features.  Savchenko \emph{et al.} \cite{savchenko2022video} proposed the novel frame-level emotion recognition algorithm which can be implemented even for video analytics on mobile devices.

\begin{figure*}[ht]
\centering
\includegraphics[scale=0.6]{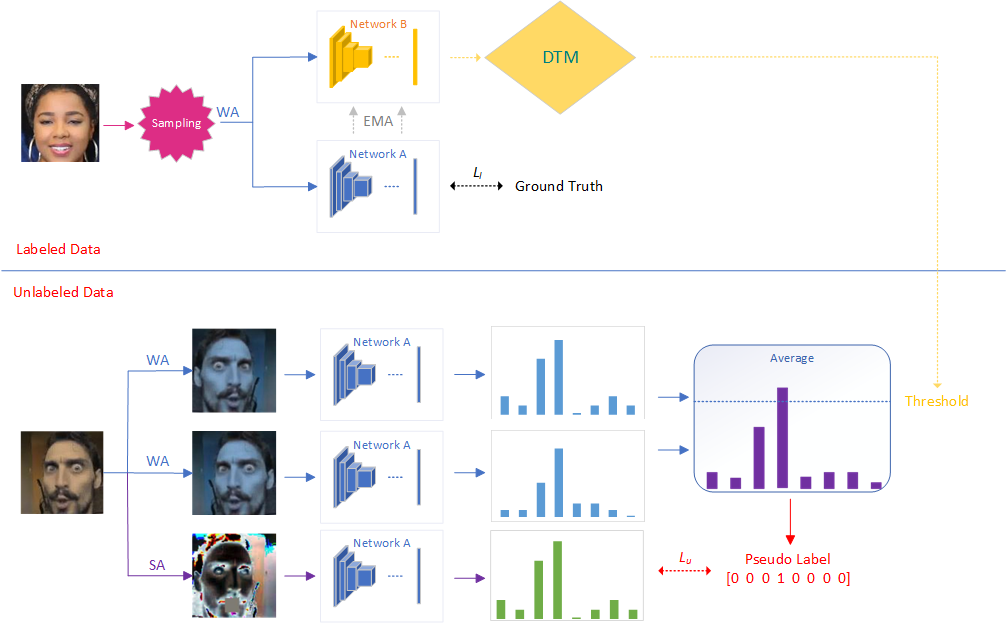}
\caption{Illustration of the framework.}
\label{fig:Pipeline}
\end{figure*}

\section{Method}
In this section, we will describe our proposed approach in detail. As shown in Figure \ref{fig:Pipeline}, the labeled data samples are weakly augmented and fed into the Network A to learn the balance expression features. The unlabeled samples with confidence greater than threshold $\tau$ is used to generate pseudo labels, which are fed into the network to learn more facial expression features. The Dynamic Threshold Module (DTM) is introduced to adjust the confidence threshold for each class dynamically in each epoch.

\subsection{Data Pre-process}
\textbf{Sampling.}
We sampled the labeled FER data to ensure that each class of expressions has the same number of samples to ensure class balance. In this way, the model learns more class-balanced features that contribute to the de-bias learning of FR data. 

\textbf{Augment.}
We conducted two kinds of data augment operations. For Weakly-Augment (WA), we mainly use horizontal flip, color jitter, etc., while  we chose RandAugment\cite{cubuk2020randaugment} as Strongly-Augment (SA) operation. The unlabeled samples remain semantic Consistency after these two data augment operations.

\subsection{Semi-Surpervised Training}
For the labeled data $(x_l, y_l)$, we sampled them and then apply weak data augmentation to obtain a probability distribution $p_l$ predicted by neural network A:
\begin{equation}
    p_l = Network_A(\emph{WA}(sampling(x_l)); \theta_A)
\end{equation}
For unlabeled data $x_u$, we first generate the weakly-augmented samples \begin{math} x_u^w \end{math} and strongly-argumented samples \begin{math} x_u^s \end{math}, then we utilize the network A to extract features and probability distributions:
\begin{equation}
    P_u^w = Network_A(\emph{WA}(x_u); \theta_A)
\end{equation}
\begin{equation}
    P_u^s = Network_A(\emph{SA}(x_u); \theta_A)
\end{equation}
Then we obtain the average probability distribution $\hat{P}_u$:
\begin{equation}
   \widetilde{P}_u = \frac{1}{2}(P_u^w + P_u^s) 
\end{equation}
If $\emph{argmax} (\widetilde{P}_u) $ is greater than the confidence threshold $\tau$, we get the pseudo label:
\begin{equation}
   \hat{y}_u = \emph{argmax}(\widetilde{P}_u)
\end{equation}

\subsection{Dynamic Threshold Module}
Fixed threshold limited the modeling ability in the early stage. Inspired by \cite{li2022towards}, we introduced a Dynamic Threshold Module (DAM) to adjust the threshold $\tau$ during different training stages. 
To enhance the robustness of the model, the network B is obtained from network A using the exponential moving average (EMA) technique with a decay rate of 0.999. We utilize the network B to extract features and probability distributions of the labeled data ($(x_l, y_l)$):
\begin{equation}
    \check{P}_l = Network_B(x_l; \theta_B)
\end{equation}
Then, we calculating the average confidence score of all correctly predicted samples in the labeled data for each class:

\begin{equation}
   \tau_c = \frac{1}{N_c} \sum_{i=1}^{N_c} p_i^c
\end{equation}

where $N_c$ is the total number of correctly predicted samples of $c$-th class, $p_i^c$ is the predicted confidence score of the correct class $c$ for the $i$-th sample, and $\tau_c$ is the threshold of $c^th$ class.

As the number of training epochs increases, the discriminative ability of the model for the trained data significantly increases. Therefore, we perform a weighted average on the threshold to prevent the threshold from increasing too quickly. Therefore, the final confidence threshold $\tau_c^t$ for class i at each epoch is:
\begin{equation}
    \tau_c^t = \mu\tau_c^{t-1} + (1-\mu)\tau_c
\end{equation}
where $\mu$ is a hyper-parameter.

\subsection{Loss Function}
We employ the cross-entropy loss\cite{chen2012cross,zhang2018generalized,ho2019real} function as the objective function for training our model.
\begin{equation}
   L_{CE} = -\sum_{i=1}^{8}y_i\log(\hat{y}_i)
\end{equation}
where $y_i$ represents the label for the i-th class, and $\hat{y}_i$ represents the predicted probability of the i-th class. 

For the labeled data, the objective function can be expressed as:
\begin{equation}
    L_l = CE(y_l, P_l)
\end{equation}

For the unlabeled data, the objective function can be expressed as:
\begin{equation}
     L_u = CE(\hat{y}_u, \widetilde{P}_u)
\end{equation}

The whole network minimizes the following loss function:
\begin{equation}
    L_{total} = \lambda_1L_l+\lambda_2L_u
\end{equation}
$\lambda_1$, $\lambda_2$ are hyper-parameters to balance each term’s intensity.

\subsection{Post-Process}
Since Aff-Wild2 \cite{zafeiriou2017aff,kollias2019deep, kollias2019face, kollias2019expression, kollias2020analysing,kollias2021affect,kollias2021analysing,kollias2021distribution,kollias2022abaw, kollias2023abaw,kollias2023abaw2}  dataset is derived from all frames of the videos, and an expression takes some time to be generated. So an obvious conclusion is that there will not be a rapid change of expressions within a few adjacent frames. So we set a sliding window to post-process the prediction results for the purpose of smoothing the prediction labels. We first count the number of all predicted labels within a window, and then consider the expression label with the most occurrences as the expression recognition result for all images within this window. Finally, we achieve the predicted expression smoothing for the whole dataset by means of a sliding window.

\begin{table*}
  \centering
  \caption{Experimental results on the official validation set}
  \label{tab:table1}
  \begin{tabular}{ccccccc}
    \toprule
    \textbf{method} & \textbf{Aff-Wild2} & \textbf{AffectNet and ExpW} & \textbf{MS1MV2} & \textbf{Post-Process} & \textbf{$F_1$ Score (\%)} \\ 
    \midrule
	baseline & $\checkmark$ & &  &  & 23.00 \\
	

	SSL + DTM & $\checkmark$ & $\checkmark$  & & & 35.81 \\
	
	SSL + DTM & $\checkmark$ & $\checkmark$ & $\checkmark$ & & 39.03\\

    SSL + DTM & $\checkmark$ & $\checkmark$ & $\checkmark$ & $\checkmark$ &    \textbf{40.05} \\
    
    \bottomrule
  \end{tabular}
\end{table*}

\section{Experiment}
In this section, we will provide a detailed description of the used datasets, the experiment setup, and the experimental results.
\subsection{Datasets}
\textbf{FER Datasets.}
The 3rd Workshop and Competition on Affective Behavior Analysis in-the-wild provides the Aff-wild2\cite{zafeiriou2017aff,kollias2019deep, kollias2019face, kollias2019expression, kollias2020analysing,kollias2021affect,kollias2021analysing,kollias2021distribution,kollias2022abaw, kollias2023abaw,kollias2023abaw2}  database as the official datasets. For EXPR Classification Challenge, This database is audiovisual (A/V) and in total consists of 548 videos of around 2.7M frames that are annotated in terms of the 6 basic expressions (i.e., anger, disgust, fear, happiness, sadness, surprise), plus the neutral state, plus a category 'other' that denotes expressions/affective states other than the 6 basic ones. In addition, we used external facial expression databases, such as AffectNet\cite{mollahosseini2017affectnet} and ExpW\cite{zhang2018facial}. AffectNet contains about 1M facial images collected from the Internet, it provides Eleven emotion and non-emotion categorical labels (Neutral, Happiness, Sadness, Surprise, Fear, Disgust, Anger, Contempt, None, Uncertain, No-Face) and we only used the first 7 categories of images. The Expression in-the-Wild Database (ExpW) contains 91,793 faces downloaded using Google image search. Each of the face images was manually annotated as one of the seven basic expression categories. In our paper, we obtained 8,000 labeled images for each category from the Aff-Wild2 dataset through uniform sampling. Additionally, to increase the diversity of our data and improve the generalization of our model, we randomly sampled 8,000 images for each category from the merged dataset of AffecNet and ExpW. For the "other" category, we used the images from the Aff-Wild2 dataset since there were no such images in the other two datasets. Finally, we included the remaining images of these datasets as the unlabeled samples.

\textbf{FR Datasets.}
For Face Recognition Datasets, We use MS1MV2\cite{huang2008labeled} as the unlabeled data. The MS1MV2 dataset is a semi-automatic refined version of the MS-Celeb-1M dataset\cite{guo2016ms} proposed by ArcFace\cite{deng2019arcface}, which includes 85k ids and 5.8m images. The unlabeled data used in our experiments consists of a subset of the InsightFace\cite{insightface} MS1MV2 dataset, obtained by uniformly sampling 1/3 of its images. This subset comprises a total of 1.94 million images.

\subsection{Setup}
All training face images are resized to 224$\times$224 pixels, Our proposed method is implemented with the PyTorch toolbox on eight NVIDIA Tesla V100 GPUs. By default, we use Efficient-B7\cite{tan2019efficientnet} as the backbone. Adam optimizer \cite{kingma2014adam} is used with the fixed learning rate of 5 $\times$ 10$^{-4}$. The batch size of labeled and unlabeled data is 32 and we conducted 1000 steps for a epoch to training the model. The hyper-parameters $\mu$, $\lambda_1$, $\lambda_2$ are set as 0.9, 1 and 0.8, respectively

\subsection{Metrics}
According to the requirements of the competition, the evaluation metrics we use is the macro
\begin{math} F_1 \end{math} score, which is not affected by the class frequency and is more suitable for the imbalanced class distribution . 
The $c$-th class $F_1^c$ can be calculated as fallows:
\begin{equation}
  F_1^c = \frac{2\times Precision\times Recall}{Precision + Recall}
\end{equation}
\begin{equation}
  F_1 = \frac{1}{N}\sum_{c=1}^{N} F_1^c
\end{equation}
where \begin{math}N \end{math} represents the number of classes and \begin{math}c \end{math} means $c$-th class. Final Macro \begin{math} F_1 \end{math} score is calculated by
computing the average score of \begin{math} F_1^c \end{math}.

\subsection{Results}
Partial experimental results are shown in Table \ref{tab:table1}. The baseline network is a pre-trained VGGFACE \cite{parkhi2015deep} (with fixed convolutional weights) and its performance on the official validation set is 0.23. By using our proposed dynamic threshold module with self-supervised learning on the Aff-Wild2, AffectNet, and ExpW datasets, we achieved 35.81\% macro $F_1$ score on the official validation set, which is a 12.81\% improvement over the baseline. We achieved this significant improvement with less than 10\% labeled FER data. To further demonstrate the effectiveness of our method on unlabeled data, we incorporated the FR dataset MS1MV2 and achieved a 3.22\% improvement, the macro $F_1$ score ranged from 35.81\% to 39.03\%. This demonstrates that our method can effectively utilize unlabeled face data to improve the performance of facial expression recognition. Finally, we applied a post-processing step with a window size of 10 on the predicted results, raising the final macro $F_1$ score to 40.05\%.

\section{Conclusion}
In this paper, we provide a semi-supervised learning approach to improve the performance of face expression recognition task using unlabeled face data, and also designs a dynamic threshold module to leverage confidence thresholds for different training stages and different expression categories to generate more accurate pseudo-labels and alleviate the dataset imbalance problem. The method has shown very good results in ABAW5 EXPR task.

\section*{Acknowledgement}
Natural Science Foundation of China (62276242), CAAI-Huawei MindSpore Open Fund (CAAIXSJLJJ-2021-016B, CAAIXSJLJJ-2022-001A), Anhui Province Key Research and Development Program (202104a05020007), USTC-IAT Application Sci. \& Tech. Achievement Cultivation Program (JL06521001Y), Sci.\&Tech. Innovation Special Zone (20-163-14-LZ-001-004-01).

{\small
\bibliographystyle{ieee_fullname}
\bibliography{egbib}
}

\end{document}